\documentclass[lettersize,journal]{IEEEtran}
\usepackage{amsmath,amsfonts}
\usepackage{algorithmic}
\usepackage{algorithm}
\usepackage{array}
\usepackage[caption=false,font=normalsize,labelfont=sf,textfont=sf]{subfig}
\usepackage{textcomp}
\usepackage{stfloats}
\usepackage{url}
\usepackage{verbatim}
\usepackage{graphicx}
\usepackage{cite}
\usepackage{multirow}
\usepackage{xcolor}
\usepackage[table]{xcolor}
\usepackage{booktabs} 
\usepackage[normalem]{ulem}

\definecolor{tab:blue}{HTML}{1F77B4}
\definecolor{tab:orange}{HTML}{FF7F0E}
\definecolor{tab:green}{HTML}{2CA02C}
\definecolor{tab:red}{HTML}{D62728}
\definecolor{tab:purple}{HTML}{9467BD}
\definecolor{tab:brown}{HTML}{8C564B}
\definecolor{tab:pink}{HTML}{E377C2}
\definecolor{tab:gray}{HTML}{7F7F7F}
\definecolor{tab:olive}{HTML}{BCBD22}
\definecolor{tab:cyan}{HTML}{17BECF}

\useunder{\uline}{\ul}{}

\usepackage{pifont}
\newcommand{\cmark}{\ding{51}}%
\newcommand{\xmark}{\ding{55}}%
\usepackage{caption}

\begin{document}

\title{Learning a Unified Latent Space for Cross-Embodiment Robot Control}

\author{Yashuai Yan$^{1}$, Dongheui Lee$^{1,2}$
\thanks{$^{1}$Yashuai Yan and Dongheui Lee are with Autonomous Systems Lab, Technische Universität Wien (TU Wien), Vienna, Austria (e-mail: \texttt{\{yashuai.yan, dongheui.lee\}@tuwien.ac.at}).\\ $^{2}$Dongheui Lee is also with the Institute of Robotics and Mechatronics (DLR), German Aerospace Center, Wessling, Germany.}
}

\markboth{Journal of IEEE Robotics \& Automation Magazine, Special Issue on Embodied AI}%
{Shell \MakeLowercase{\textit{et al.}}: A Sample Article Using IEEEtran.cls for IEEE Journals}


\maketitle


\begin{abstract}
We present a scalable framework for cross-embodiment humanoid robot control by learning a shared latent representation that unifies motion across humans and diverse humanoid platforms, including single-arm, dual-arm, and legged humanoid robots. Our method proceeds in two stages: first, we construct a decoupled latent space that captures localized motion patterns across different body parts using contrastive learning, enabling accurate and flexible motion retargeting even across robots with diverse morphologies. To enhance alignment between embodiments, we introduce tailored similarity metrics that combine joint rotation and end-effector positioning for critical segments, such as arms. Then, we train a goal-conditioned control policy directly within this latent space using only human data. Leveraging a conditional variational autoencoder, our policy learns to predict latent space displacements guided by intended goal directions. We show that the trained policy can be directly deployed on multiple robots without any adaptation. Furthermore, our method supports the efficient addition of new robots to the latent space by learning only a lightweight, robot-specific embedding layer. The learned latent policies can also be directly applied to the new robots. Experimental results demonstrate that our approach enables robust, scalable, and embodiment-agnostic robot control across a wide range of humanoid platforms.
\end{abstract}


\section{Introduction}
\IEEEPARstart{C}{ontrolling} a diverse range of humanoid platforms through a unified interface remains a long-standing ambition in robotics. Achieving this goal, however, is highly challenging due to the vast variability in robot morphology, degrees of freedom, and kinematic constraints. A key requirement for scalable generalization is to develop control policies that can seamlessly transfer across embodiments without retraining or platform-specific finetuning. In this work, we address the problem of cross-embodiment humanoid robot control by learning a unified shared latent space. Control policies trained in the latent space can be deployed on any robot that is encoded in this space.

Recent advancements in robot learning, particularly in cross-domain imitation learning \cite{pmlr-v119-kim20c}, have enabled more flexible and adaptable robot control across different embodiments. A common approach to bridging the embodiment gap is to learn task-relevant, domain-invariant representations \cite{9811668, franzmeyer2022learn, pmlr-v164-zakka22a}. For example, to enable robots to learn manipulation skills from human demonstrations, prior work \cite{pertsch2022star, 10610084, pmlr-v229-xu23a} aligns cross-domain skill representations between human videos and robot demonstrations. However, these methods often require a substantial amount of paired demonstration data from both domains, limiting scalability and hindering transferability to new, unseen platforms—an increasingly important consideration given the rapid diversification of robotic morphologies.

Beyond task-relevant feature extraction, another promising strategy for bridging the embodiment gap is to construct a shared latent space between embodiments, where semantically similar actions are mapped to proximate points regardless of the originating domain. Early efforts in this direction \cite{rt_pair_data_1, rt_pair_data_2, rt_shared_space_humanoid} focused on learning shared representations for translating motions from humans to robots or animated characters, but relied heavily on manually collected paired datasets. To overcome this data bottleneck, Choi et al. \cite{rt_selfsupervised_hyemin} proposed a self-supervised paired data generation technique, automating the process of building correspondences between domains. More recently, Yan et al. \cite{imitationnet} introduced a contrastive learning approach to discover a shared latent space, enhancing both the expressiveness and smoothness of motion retargeting between humans and robots. However, the prior work addressed the retargeting from humans to a specific robot embodiment. Our primary goal is to learn a unified latent space across diverse embodiments, which later allows unified policies to control multiple robots.

In this work, we tackle the problem of cross-embodiment robot control, where the objective is to develop a single control framework capable of operating across diverse robotic embodiments. Specifically, we aim to achieve two key capabilities: (1) learning a shared latent representation that captures motion semantics across diverse embodiments, and (2) training a goal-conditioned control policy within this shared space that can be directly deployed to all robots that are encoded in the shared space.

To this end, we propose a two-stage approach. In the first stage, we learn a unified latent space that aligns motions from varied embodiments into a common representation. Building on insights from contrastive learning and motion retargeting, we introduce a decoupled latent structure that models different body parts independently (arms, legs, trunk). This modular design allows for finer-grained alignment across embodiments with asymmetric or partial limb structures and enables the use of customized similarity metrics, such as combining joint rotation and end-effector positioning for arm movements. Our method not only unifies diverse embodiments into a shared latent space 
but also improves the cross-embodiment motion retargeting performance.

In the second stage, we leverage the learned latent space to train a goal-conditioned motion generation policy based purely on human motions. Using a conditional variational autoencoder (c-VAE) framework, we model motion generation as predicting latent displacements guided by intention features derived from end-effector velocities toward user-specified goals. By operating directly in the latent space, the policy achieves embodiment-agnostic control, allowing motions trained on human demonstrations to be seamlessly executed across multiple robot platforms. Importantly, the modularity of our system enables efficient scaling to new robots: integrating a new embodiment requires only learning a lightweight, robot-specific embedding layer, without re-training the entire model.

The proposed approach has the following contributions.
\begin{itemize}
    \item a decoupled latent space architecture that enables fine-grained motion alignment across a wide range of robot morphologies.

    \item a modular contrastive learning framework with tailored similarity metrics, capable of jointly learning motion representations from both human and multiple robot embodiments.

    \item a goal-conditioned control policy that operates directly within the shared latent space, allowing accurate multi-robot control without any additional fine-tuning.

    \item a scalable framework by adding new robots through lightweight embedding training, enabling efficient deployment in multi-robot systems
\end{itemize}

\section{Methodology}
\begin{figure*}[]
    \centering
    \includegraphics[width=0.9\textwidth]{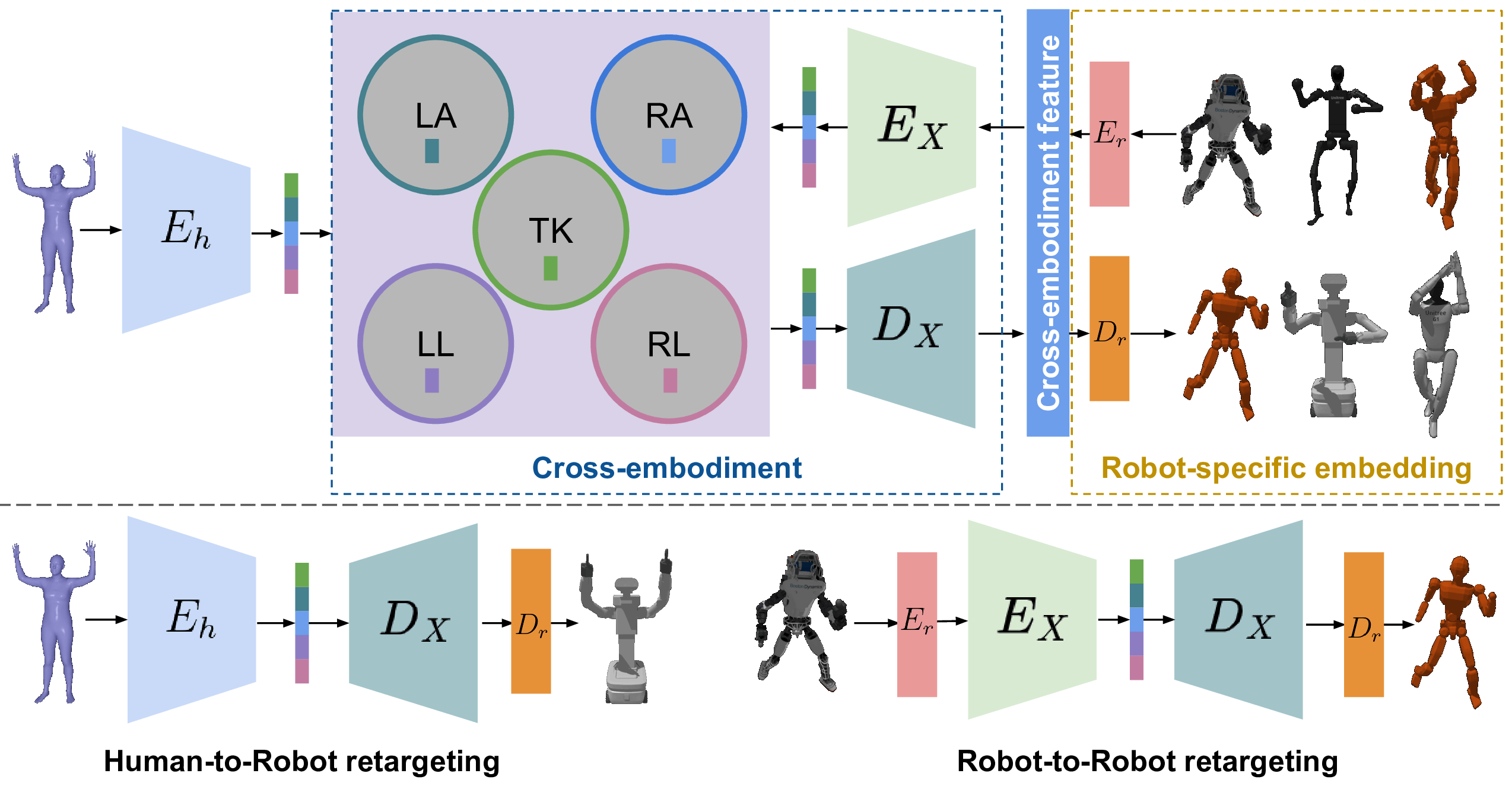}
    \caption{\textbf{Learning a Unified Latent Representation.} Our architecture learns a shared latent space that unifies motion representations across diverse embodiments, including humans and various robots. To accurately model local motion patterns, we decouple the latent space into five subspaces corresponding to distinct body segments: left arm (LA), right arm (RA), trunk (TK), left leg (LL), and right leg (RL). The model comprises a human encoder ($E_h$), a cross-embodiment encoder ($E_X$), and a cross-embodiment decoder ($D_X$). To accommodate differences in pose dimensionality across robot platforms, each robot is assigned a learnable, robot-specific embedding layer $E_r$, which projects the raw pose representation into a shared cross-embodiment feature space. Conversely, $D_r$ is the inverse mapping back to the original pose space.}
    \label{fig:modeloverview}
\end{figure*}

In this section, we address the cross-embodiment robot control as a two-stage problem. Our method first learns a unified latent space over different robots. Then, we formulate multi-robot control as controlling the shared latent space. An overview of the two stages is illustrated in Fig. \ref{fig:modeloverview} and Fig. \ref{fig:cvae}, respectively.

\subsection{Problem Formulation}
To enable control across a diverse set of robotic embodiments within a multi-robot system, our approach comprises two main components: (1) learning a shared representation across all robots, and (2) training a control policy within this shared space.

We tackle the challenge of learning a shared latent space by enabling robots to retarget motions from other embodiments within this latent representation. To formalize the cross-embodiment motion retargeting problem, we represent a pose of embodiment $A$ as $\mathbf{x}_{A} \in \mathbb{R}^{J_A \times n_A}$, where $A$ denotes a specific embodiment (e.g., a human or a robot), $J_A$ is the number of joints in embodiment $A$, and $n_A$ is the dimensionality of its joint representation. In our setup, human joints are represented using quaternions ($n=4$) while robot joints use scalar joint angles ($n=1$). 
A motion for Embodiment $A$ is denoted as $\mathbf{X}_A^{1:T} = [\mathbf{x}_{A}^{1}, \mathbf{x}_{A}^{2}, \cdots, \mathbf{x}_{A}^{T}]$. The goal of cross-embodiment motion retargeting is to learn a mapping function $f: \mathbf{X}_A^{1:T} \rightarrow \mathbf{Y}_B^{1:T}$ that transfers a motion sequence from embodiment $A$ to a corresponding motion in embodiment $B$, such that the retargeted motion $\mathbf{Y}_B^{1:T}$ preserves the style and intent of the original motion $\mathbf{X}_A^{1:T}$.

We build on ImitationNet \cite{imitationnet}, leveraging its unsupervised, contrastive learning-based approach to learn a shared latent space. Unlike the original formulation, limited to human-robot transfer, our method generalizes this latent space to accommodate arbitrary embodiments, facilitating broader cross-platform motion retargeting.

\subsection{Segment-aware Latent Space and Similarity Metric}
\label{sec:metric}
The prior approaches \cite{rt_shared_space_humanoid, imitationnet} employ a single latent space to represent whole-body motions. However, this design can lead to ambiguous mappings between embodiments with differing morphologies. For example, the ATLAS robot possesses articulated arms, legs, and a trunk, while the H1 robot exhibits limited trunk articulation, and the TIAGO robot is restricted to arm movements only.

To address this limitation, we propose decoupling the latent space into distinct subspaces for different body segments. Specifically, we segment the body into five components: left arm (LA), right arm (RA), trunk (TK), left leg (LL), and right leg (RL). Each segment is modeled with its own dedicated latent space, allowing us to better capture localized motion features and accommodate embodiments with partial or asymmetric limb structures.

The decomposition of latent space also enables the use of customized similarity metrics for contrastive learning within each subspace. For example, end-effector position is critical in robot applications such as manipulation. In contrast, mimicking limb rotations is important for the trunk and legs to preserve visual fidelity during motion retargeting. Therefore, we design a hybrid similarity metric for the arm segments that jointly accounts for limb rotations and end-effector positions, while applying a rotation-based similarity metric to the other body segments.

We define the rotation-based similarity metric as follows:
\begin{equation}
    D_{R}(\mathbf{x}_{A}, \mathbf{x}_{B}) = \sum_{j}(1 - <q_A^j, q_B^j>^{2}),
\label{eq:rot}
\end{equation}
where $\mathbf{x}_{A}$ and $\mathbf{x}_{B}$ are poses from embodiment $A$ and $B$, respectively. The index $j$ denotes the number of joints within the body part being compared—for instance, $j=2$ for a typical arm (lower and upper joints). The term $q_A^j$ represents the quaternion of the $j$-th joint in embodiment $A$, and $<\cdot,\cdot>$ denotes the dot product between two quaternions. 

To ensure consistency across embodiments with differing arm lengths and coordinate frames, we first express each EE position relative to the shoulder frame, and then normalize it by the corresponding arm length. The EE-based similarity is computed as:
\begin{equation}
    D_{ee}(\mathbf{x}_{A}, \mathbf{x}_{B}) = ||p_A^{ee} - p_B^{ee}||_2,
\label{eq:pos}
\end{equation}
where $p_A^{ee}$ and $p_B^{ee}$ denote the normalized positions of embodiments $A$ and $B$, respectively.

For the arm latent subspaces (LA and RA), we compute the similarity as a weighted sum of rotation and EE-based distances:
\begin{equation}
\begin{aligned}
    S_k(\mathbf{x}_{A}, \mathbf{x}_{B}) = D_{R}(\mathbf{x}_{A}, \mathbf{x}_{B}) + \omega D_{ee}(\mathbf{x}_{A}, \mathbf{x}_{B}) \\
    \text{for} \quad k \in \{LA, RA\},
\end{aligned}
\label{eq:similarity}
\end{equation}
where $\omega$ is a tunable weight balancing the importance of orientation versus end-effector positioning.

For the remaining body parts (TK, LL, RL), we rely solely on the rotation-based metric:
\begin{equation}
\begin{aligned}
    S_k(\mathbf{x}_{A}, \mathbf{x}_{B}) = D_{R}(\mathbf{x}_{A}, \mathbf{x}_{B}) \\ \text{for} \quad k \in \{TK, LL, RL\}.
\end{aligned}
\end{equation}
This modular similarity framework allows us to more accurately align motions between embodiments with different capabilities, ultimately improving the quality of cross-embodiment motion retargeting.

\subsection{Learning a Unified Latent Space}
To enable motion retargeting across a diverse set of embodiments, we train a unified latent representation shared among humans and multiple robot types. This shared latent space is optimized using contrastive learning applied independently to each decoupled body-part subspace, as described in Section \ref{sec:metric}. The key idea is to embed semantically similar motions—regardless of embodiment—closely together in their respective latent subspaces.

Figure \ref{fig:modeloverview} presents an overview of the model architecture used to learn this unified latent space. The system comprises a human encoder $E_h$, a cross-embodiment encoder $E_X$, a cross-embodiment decoder $D_X$, and robot-specific embedding layers ($E_r, D_r$). Since robots may have differing numbers of joints, their raw pose vectors $\mathbf{x} \in \mathbb{R}^{J\times n}$ vary in dimensionality. To handle this variation, each robot pose is first passed through a learnable robot-specific embedding layer, which transforms it into a fixed-size high-dimensional vector. This transformation ensures consistent input dimensionality to the shared encoder. 

After embedding, robot poses are processed by the shared encoder, which outputs five latent vectors, corresponding to five latent decoupled subspaces (LA, RA, TK, LL, RL). Similarly, the human encoder network $E_h$ maps the human poses into five latent vectors, each aligned with one of the body-part-specific subspaces. $D_X$ then reconstructs the latent vectors back into a shared embedding space, which is finally projected back to robot-specific joint angles by $D_r$. 

During training, each batch of data contains pose samples from both human and robot domains. For each subspace, we apply contrastive learning using randomly sampled triplets composed of poses from any embodiment. A triplet consists of an anchor sample, a positive sample that is semantically similar, and a negative sample that is less similar. 

Let ($z_i^o, z_j^+$, $z_k^-$) represent a triplet in a given subspace, where $z_i^o$ is the anchor, $z_j^+$ is the positive sample, and $z_k$ is the negative sample.  We train the latent space using the Triplet Loss \cite{hoffer2018deep}, defined as:
\begin{multline}
    \mathcal{L}_{\text{contrastive}} = \sum_{\mathcal{S}} \sum_{(z_i^o, z_j^+, z_k^-) \in \mathcal{S}} \\
     \max(||z_i^o - z_j^+||_2 - ||z_i^o - z_k^-|| + \alpha, 0),
\end{multline}
where $\mathcal{S}$ denotes each subspace, and $\alpha$ is a margin hyperparameter that enforces a minimum separation between the positive and negative samples. Intuitively, the loss encourages the positive sample to lie closer to the anchor than the negative sample by at least $\alpha$. In our experiments, we set $\alpha = 0.05$.

To train the decoder $D_X$ to reconstruct robot motions from the latent space accurately, we apply a reconstruction loss $\mathcal{L}_{\text{rec}}$. This loss ensures that the decoded output closely matches the original input robot pose. Given an input robot pose $\mathbf{x}_A$ from embodiment $A$, and the reconstructed pose $\mathbf{\hat{x}}_A$, the reconstruction loss is defined as:
\begin{equation}
    \mathcal{L}_{\text{rec}} = ||\mathbf{x}_A - \mathbf{\hat{x}}_A||_2.
\label{eq:rec}
\end{equation}
Here, $A$ denotes any robotic embodiment. The reconstruction loss serves to regularize the latent-to-pose mapping, ensuring that the decoder maintains fidelity to the input motion during encoding and decoding.

However, for human motion data, $\mathbf{x}_H$, direct reconstruction is not possible due to the lack of paired human-to-robot data. Instead, we employ a latent consistency loss $\mathcal{L}_{\text{ltc}}$, inspired by the circular loss introduced in ImitationNet \cite{imitationnet}. This loss encourages the decoder to generate robot motions that remain faithful to the original human motion when viewed in the latent space.

Specifically, $\mathcal{L}_{\text{ltc}}$ measures the discrepancy between the original human latent representation $E_h(\mathbf{x}_H)$ and the latent representation obtained by re-encoding the decoded robot motion: $E_X(D_X(E_h(\mathbf{x}_H)))$. The loss is formally defined as:
\begin{equation}
    \mathcal{L}_{\text{ltc}} = ||E_h(\mathbf{x}_H) - E_X(D_X(E_h(\mathbf{x}_H)))||_2
\end{equation}
This latent consistency objective ensures that motions generated by decoding human embeddings can be re-encoded to yield representations close to the original human latent space. It effectively aligns the human and robot domains, thereby improving the quality of cross-embodiment motion retargeting.

Finally, to enhance the temporal consistency of retargeted motions, we introduce a temporal loss $\mathcal{L}_{\text{temporal}}$, which focuses on aligning the end-effector velocities between human and robot motions. Specifically, we consider two consecutive human poses, $\mathbf{x}_H^t$ and $\mathbf{x}_H^{t+1}$, and their corresponding retargeted robot poses, $\mathbf{x}_A^t$ and $\mathbf{x}_A^{t+1}$, where $A$ denotes a target robot embodiment.

From these consecutive frames, we compute the human hand velocity $v_H^{\text{hand}}$ and robot EE velocity $v_A^{\text{ee}}$. The temporal loss is then defined as the L2 distance between these velocity vectors:
\begin{equation}
    \mathcal{L}_{\text{temporal}} = ||v_H^{\text{hand}} - v_A^{\text{ee}}||_2
\label{eq:loss_v}
\end{equation}

To train the model end-to-end, we combine all previously defined objectives into a single weighted loss function:
\begin{equation}
    \mathcal{L}_{\text{total}} = \lambda_c \mathcal{L}_{\text{contrastive}} + \lambda_{\text{rec}} \mathcal{L}_{\text{rec}} + \lambda_{ltc} \mathcal{L}_{\text{ltc}} + \lambda_{temp} \mathcal{L}_{\text{temporal}}
\label{eq:total_loss}
\end{equation}
where we empirically set $\lambda_c=10, \lambda_{\text{rec}}=5, \lambda_{ltc}=1, \lambda_{temp}=0.1$.

By minimizing $\mathcal{L}_{\text{total}}$, the model learns a unified latent space in which motions can be accurately and smoothly retargeted across a wide range of embodiments. This shared space enables us to train control policies within the latent domain and seamlessly deploy them across multiple robots, regardless of their structural differences.

\subsection{Adding New Robots}
Our architecture is designed to facilitate efficient scalability by maintaining a shared latent space and shared network components across all embodiments. While the core encoders and decoder—$E_h, E_X$, and $ D_X$—are jointly trained across multiple robots and humans, each robot retains its own learnable embedding layer to account for embodiment-specific differences.

This modular design enables seamless integration of new robots into the system. Specifically, once the shared components of the model are pre-trained, we can freeze the networks $E_h, E_X$, and $D_X$, and train only the embedding layers ($E_r, D_r$) for the new robot. This approach allows the new robot to align with the existing latent space without requiring retraining of the entire model, reducing computational cost and training time.

As a result, our framework supports scalable deployment to new robotic platforms while preserving the generalization and robustness of the unified control policy learned in the latent space.

\begin{figure}
    \centering
    \includegraphics[width=0.95\linewidth]{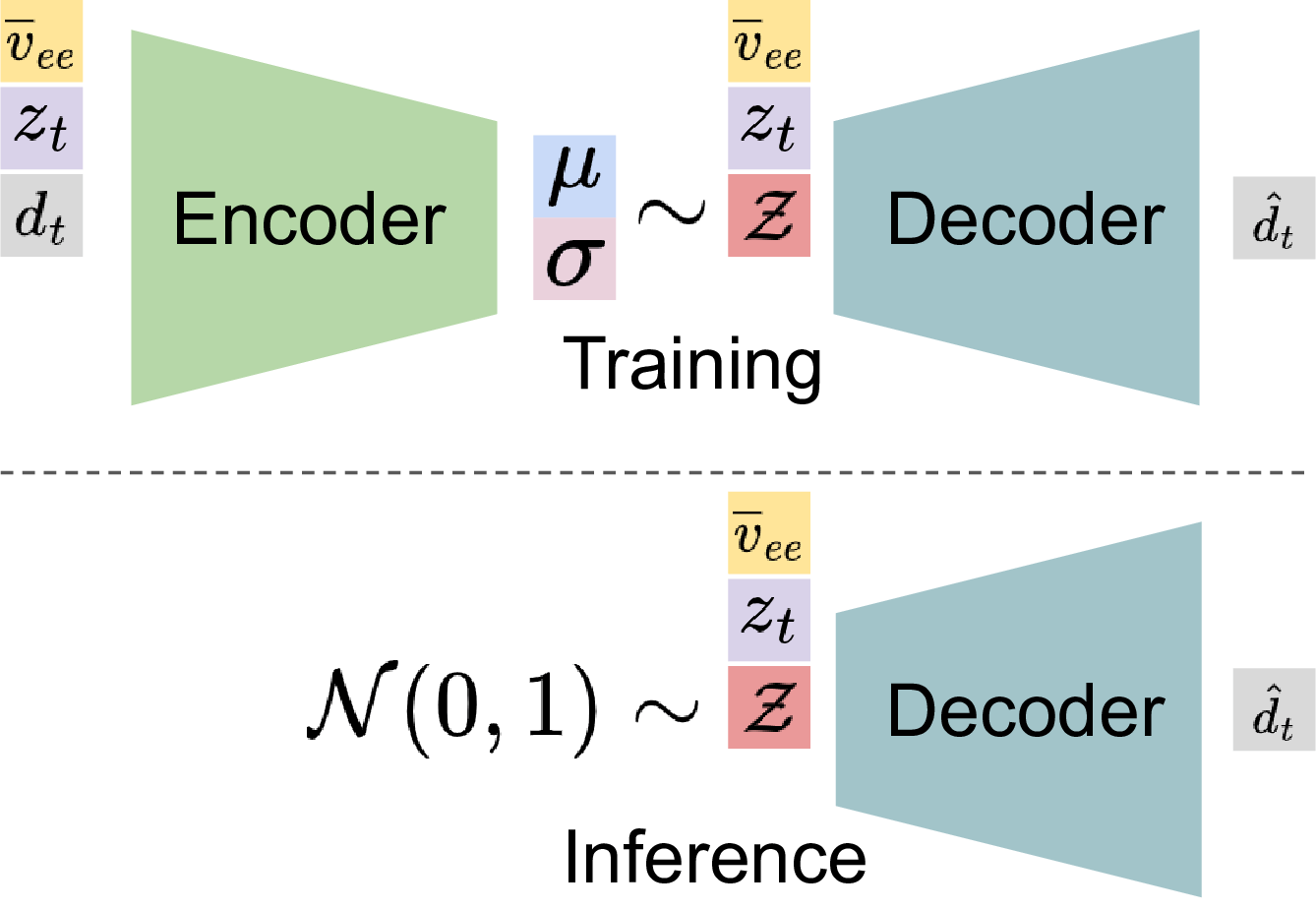}
    \caption{\textbf{Goal-conditioned Latent-Space Robot Control.} Our c-VAE framework learns to model goal-directed motion dynamics in our shared latent space purely from human demonstrations. During training, the model conditions on the current latent pose $z_t$ and the average EE velocity toward a sampled goal $\overline{v}_{ee}$ to predict the latent displacement $d_t$. At inference time, $\overline{v}_{ee}$ is derived from the current robot pose and the user-specified goal position and time horizon. The decoder then generates the latent displacement $\hat{d}_t$ using $z_t$, $\overline{v}_{ee}$, and sampled latent noise. This autoregressive process iteratively updates the latent state, enabling smooth and goal-directed motion generation across robot embodiments.}
    \label{fig:cvae}
\end{figure}
\subsection{Latent-Space Robot Control}
\label{sec:latent-space-robot-control}
With our cross-embodiment retargeting framework in Fig.~\ref{fig:modeloverview}, human and robot motions are projected into a unified latent space, where semantically similar movements across embodiments are encoded close together. This shared representation enables a high-level abstraction of robot control: rather than specifying joint commands, robot behavior can be modulated by traversing this latent space.

To exploit this capability, we propose a goal-conditioned motion generation policy that operates directly in the latent space and is trained exclusively on human data. Crucially, this policy can be deployed to multiple robot platforms without any further fine-tuning. We focus on a practical and widely applicable control objective in robotics—generating arm movements that guide the robot’s EE to reach a desired goal position.

We model this task using the c-VAE architecture, where the motion generation is guided by control signals. During training, we utilize human hand positions to extract the goal signal. Due to the modular structure of our decoupled latent space (see Sec.~\ref{sec:metric}), the policy targets only the relevant subspace (e.g., LA or RA), allowing efficient and focused learning for arm control.

Training proceeds as follows: from a human motion sequence $\mathbf{X}_H^{1:T}$, we sample three frames: $\mathbf{x}_H^t, \mathbf{x}_H^{t+1}$ and  a future goal frame $\mathbf{x}_H^{t_G}$. Hand positions $p_{\text{hand}}^{t}$ and $p_{\text{hand}}^{t_G}$ are extracted from these frames, and the average velocity vector toward the goal is computed as $\overline{v}_{\text{ee}} = \frac{p_{\text{hand}}^{t_G} - p_{\text{hand}}^t}{t_G - t}.$
$\overline{v}_{\text{ee}}$ serves as an intention signal, providing directional guidance for generating future motion. Concurrently, the human encoder $E_h$ encodes the poses $\mathbf{x}_H^t$ and $\mathbf{x}_H^{t+1}$ into latent representation $z_t$ and $z_{t+1}$, respectively.

Rather than reconstructing absolute poses, our model learns to predict the latent displacement vector $d_t = z_{t+1} - z_t$, which captures motion dynamics between consecutive frames. Predicting these deltas also benefits from the inherent inductive biases of sequential datasets, which contribute to more sample-efficient learning and improved generalizability, as noted in prior work \cite{diomataris2024wandr, ling2020character, rempe2021humor}. The design of architecture (in Fig.~\ref{fig:cvae}), conditioning on $z_t$ and $\overline{v}_{ee}$, ensures that motion generation is informed both by the current pose and by the goal-directed intention.

The overall training objective for the c-VAE consists of a reconstruction loss and a KL divergence regularization: 
\begin{equation} 
\mathcal{L}_{\text{cvae}} = \mathcal{L}_{\text{reconstruction}} + \lambda_{\text{KL}} \mathcal{L}_{\text{KL}}. 
\end{equation}

The reconstruction loss, $\mathcal{L}_{\text{reconstruction}} = ||d_t - \hat{d}_t||_2^2$, encourages the faithful reproducing of human motions.

The KL divergence term, $\mathcal{L}_{\text{KL}} = D_{\text{KL}}(\mathcal{N}(0, I) || \mathcal{N}(\mu, \sigma))$, aligns the encoder's output distribution with a standard Gaussian prior. $\mu$ and $\sigma$ are the predicted mean and variance of the latent distribution. We set $\lambda_{KL} = 10^{-4}$ to ensure that the regularization term supports, but does not dominate the reconstruction objective.

At inference time, we deploy the learned policy to generate robot motions toward a user-specified goal. The process begins with the user providing a target EE position $p_{ee}^T$ and a time horizon $T$. Given the current EE position $p_{ee}^t$ at time step $t$, we compute the goal-directed velocity vector as $\overline{v}_{ee} = \frac{p_{ee}^T - p_{ee}^t}{T - t}$.

Simultaneously, the current robot pose is encoded into the latent space as $z_t$. The decoder, conditioned on $z_t$ and the intention vector $\overline{v}_{ee}$, along with a sampled latent noise, predicts the latent displacement $\hat{d}_t$ that represents the transition to the next state. The next latent representation is then computed as $z_{t+1} = z_t + \hat{d}_t$.

Importantly, at each step, the intention vector $\overline{v}_{ee}$ is dynamically updated based on the predicted EE position and the original goal, enabling the policy to adapt to real-time changes and maintain goal-oriented motion. This process is executed in an autoregressive fashion, iteratively generating a sequence of latent states that are subsequently decoded into robot joint configurations.

\section{Experiments}
This section is organized as follows: We begin by outlining the implementation details to support reproducibility. Next, we introduce the baseline methods. Finally, we present a comprehensive evaluation of our approach on multiple robots with different morphologies.

\subsection{Technical Implementation}
\subsubsection{Learning a Unified Latent Space}
The two encoders ($E_h$ and $E_X$) and the decoder ($D_X$) in our architecture are implemented using multilayer perceptrons (MLPs). Each MLP consists of 8 fully connected layers, with 256 neurons per layer. We use Exponential Linear Units (ELU) \cite{clevert2016fast} as the activation function for intermediate layers and apply a Tanh activation at the output layer to constrain the output range. Each latent subspace is 16-dimensional, with values bounded between -1 and 1.

To handle embodiment-specific differences in pose dimensionality, each robot is equipped with a learnable embedding layer $E_r$ that projects its joint space into a 1024-dimensional feature space. An inverse embedding layer $D_r$ is used to map this shared representation back into the robot-specific joint space, enabling the reconstruction of the original pose.

\subsubsection{Latent-Space Robot Control}
Similarly, the encoder and decoder networks in the c-VAE are also implemented as MLPs, each comprising 8 linear layers. Every layer is followed by an ELU activation, except for the output layer, which uses no activation function to allow unbounded outputs. The encoder outputs a 32-dimensional latent distribution, representing the mean and variance parameters of a Gaussian distribution from which latent variables are sampled during training. 

\subsubsection{Training Details}
We developed and trained our method using the PyTorch framework. For optimization, we use Adam \cite{kingma2014adam} with a constant learning rate of $10^{-3}$. Our models are trained with a batch size of $10^5$ on an NVIDIA A4000 GPU.

\subsection{Datasets}
We use the HumanML3D dataset \cite{HumanML3D}, which contains 29,224 diverse human motion sequences comprising over 4 million human poses. Notably, our approach does not require any robot data collection. Instead, we sample robot joint configurations uniformly at random from their respective joint spaces during training and compute the corresponding robot poses using forward kinematics (FK). For FK computation, we employ PyTorch-Kinematics \cite{Zhong_PyTorch_Kinematics_2024}, which efficiently parallelizes FK calculations on the GPU using only robot URDFs. This design enables the entire training pipeline—from data sampling to neural network updates—to be executed entirely on GPUs, thereby improving training efficiency. At each training step, over $10^5$ new robot poses are sampled and immediately discarded after updating the networks, ensuring diverse and unbiased exposure without requiring dataset storage. Over the course of training, this results in billions of robot poses per embodiment. This large-scale synthetic sampling enables our model to explore the robot’s kinematic space comprehensively and supports the formation of a smooth and expressive latent space.

\subsection{Baselines}
To assess the performance of our framework, we evaluate it across three key settings: human-to-robot motion retargeting, cross-embodiment generalization, and robot end-effector control. Each baseline is selected to highlight specific aspects of our method’s capability—representational efficiency, embodiment scalability, and control precision.

\subsubsection{Human-to-Robot Motion Retargeting}
We first compare our method against ImitationNet \cite{imitationnet}, a deep-learning-based approach originally designed to retarget motions from human demonstrations to a dual-arm TIAGo++ robot. Unlike our unified latent space that supports multiple embodiments, ImitationNet learns a shared space from humans for one single robot. For a fair evaluation, we train individual ImitationNet models for each target robot, such as human-to-JVRC and human-to-H1. This comparison highlights the scalability and generalization of our shared latent representation, which enables joint learning across all robots in a single model.

\subsubsection{Cross-Embodiment Motion Retargeting}
To evaluate the effect of our decoupled latent space, we compare it against a monolithic, whole-body latent representation trained under an otherwise identical setup. We hypothesize that separating the latent space into semantically aligned subspaces (e.g., arms, legs, trunk) provides better modularity and embodiment-
invariance—critical when mapping motions between diverse
morphologies. By toggling between the decoupled and coupled
variants of our framework, we isolate and quantify the impact
of this architectural choice.


\subsection{Quantitative and Qualitative Evaluation}
To evaluate our models on diverse and unseen motions, we split the HumanML3D dataset into $80\%$ of training data and $20\%$ of testing data.

\subsubsection{Metrics}
To quantitatively evaluate motion retargeting quality and goal-reaching performance, we define the following metrics:

\begin{itemize} 
\item \textbf{Rotation Similarity (RS)} measures the rotational consistency of body limb orientations across different embodiments. Specifically, it computes the average angular distance between corresponding joints as defined in Eq.~\ref{eq:rot}.

\item \textbf{Normalized Distance Similarity (NDS)} evaluates the spatial alignment of end-effector (or hand) positions after normalization. To account for scale differences across embodiments, all positions are transformed into the shoulder-local coordinate frame and normalized by arm length, as detailed in Sec.~\ref{sec:metric}. The formal definition is given in Eq.~\ref{eq:pos}.

\item \textbf{Normalized Velocity Similarity (NVS)} assesses the similarity of end-effector velocities between retargeted and reference motions. This metric is crucial for validating the consistency of latent dynamics across embodiments, as it reflects the correspondence between motion intent and execution. NVS is computed as in Eq.~\ref{eq:loss_v}.

\item \textbf{Distance to Goal (DTG)} measures the absolute position distance of the end-effector in goal-conditioned tasks. It is defined as the Euclidean distance between the predicted end-effector position at the final frame and the user-specified goal location.
\end{itemize}
\subsubsection{Motion Retargeting}
\begin{table*}[]
\centering
\resizebox{\textwidth}{!}{%
\begin{tabular}{ccccccccccc}
\multicolumn{1}{l}{} &
  \multicolumn{1}{l}{} &
  \multicolumn{3}{c}{Rotation Similarity (in degree)} &
  \multicolumn{3}{c}{Normalized Distance Similarity} &
  \multicolumn{3}{c}{Normalized Velocity Similarity} \\ \cmidrule(lr){3-5} \cmidrule(lr){6-8} \cmidrule(lr){9-11}
source &
  target &
  ImitationNet &  \begin{tabular}[c]{@{}c@{}}ours \\ (coupled)\end{tabular} &  \begin{tabular}[c]{@{}c@{}}ours\\ (decoupled)\end{tabular} &
  ImitationNet & \begin{tabular}[c]{@{}c@{}}ours \\ (coupled)\end{tabular} & \begin{tabular}[c]{@{}c@{}}ours\\ (decoupled)\end{tabular} &
  ImitationNet & \begin{tabular}[c]{@{}c@{}}ours \\ (coupled)\end{tabular} & \begin{tabular}[c]{@{}c@{}}ours\\ (decoupled)\end{tabular} \\ \cmidrule(lr){1-2} \cmidrule(lr){3-5} \cmidrule(lr){6-8} \cmidrule(lr){9-11}
\multirow{4}{*}{Human} & TIAGo++ & 
  \textbf{0.7183} & 
  4.2622 &
  {\ul 3.8293} &
  0.1325 &
  {\ul 0.0492} &
  \textbf{0.0401} &
  0.3762 &
  {\ul 0.1252} &
  \textbf{0.1071} \\
 &
  H1 &
  \textbf{0.6483} &
  2.1268 &
  {\ul 1.0947} &
  0.1081 &
  {\ul 0.0353} &
  \textbf{0.0263} &
  0.2881 &
  {\ul 0.1116} &
  \textbf{0.0962} \\
 &
  NAO &
  \textbf{0.6685} &
  4.2371 &
  {\ul 2.7097} &
  0.1596 &
  {\ul 0.0635} &
  \textbf{0.0566} &
  0.3682 &
  {\ul 0.1604} &
  \textbf{0.1448} \\
 &
  JVRC &
  \textbf{0.5792} &
  1.9812 &
  {\ul 1.4631} &
  0.1006 &
  {\ul 0.0289} &
  \textbf{0.0288} &
  0.2383 &
  {\ul 0.1001} &
  \textbf{0.0862} \\  \midrule
JVRC &
  H1 &
  - &
  0.6962 &
  \textbf{0.4699} &
  - &
  0.0232 &
  \textbf{0.0206} &
  - &
  \textbf{0.0822} &
  0.0869 \\  
H1 &
  NAO &
  - &
  2.7630 &
  \textbf{1.8093} &
  - &
  \textbf{0.0502} &
  0.0511 &
  - &
 0.1316 &
  \textbf{0.1300} \\
TIAGo++ &
  JVRC &
  - &
  3.3615 &
  \textbf{2.2428} &
  - &
  0.0402 &
  \textbf{0.0360} &
  - &
  0.1058 &
  \textbf{0.1040} \\ \midrule
\end{tabular}%
}
\caption{\textbf{Motion Retargeting.} Comparison of different imitation models. Notice that ImitationNet trains an individual model for each human-to-robot mapping, while our method trains a unified model across all embodiments. We also conduct ablation study on the latent space decomposition for cross-embodiment retargeting.}
\label{tab:retargeting}
\end{table*}

We evaluate our method and the baselines on the testing dataset using the defined metrics. Our focus is on assessing cross-embodiment motion retargeting performance across four distinct robot platforms: TIAGo++, H1, NAO, and JVRC. These robots span a range of morphologies, from mobile-base bimanual arms (TIAGo++) to fully legged humanoids (JVRC), introducing varying kinematic constraints and control challenges. Detailed specifications of each robot are provided in Appendix~\ref{appendix:robots}.

For the baseline ImitationNet~\cite{imitationnet}, a separate model is trained for each robot, while our method trains a single unified model across human and all robot embodiments. Table~\ref{tab:retargeting} presents a quantitative comparison of our approach against the baselines. Our method significantly outperforms existing techniques in terms of end-effector position and velocity accuracy, both of which are critical for enabling precise goal-conditioned control policies in the learned latent space.

However, we observe a decline in rotational accuracy when using a shared latent space across all embodiments. This degradation arises due to intrinsic differences in robot kinematics: for instance, TIAGO is restricted to arm motions, whereas H1 and NAO possess legged locomotion but have limited trunk mobility compared to humans or JVRC. To mitigate this, we adopt a decoupled latent space design, which not only recovers rotational performance comparable to ImitationNet but also yields further improvements in EE trajectory fidelity.

Furthermore, the use of a unified model across diverse embodiments enables our method to support motion transfer among different robots via the shared latent space. This capability demonstrates the potential of our approach to scale to multi-robot systems, where a single policy can be used to control multiple robots. We explore this application further in the next section.

\subsubsection{Adding New Robots}
\begin{table}[]
\centering

\begin{tabular}{lcccc}
source                 & \multicolumn{1}{l}{target} & \multicolumn{1}{l}{RS (degree)} & NDS    & NVS   \\ \midrule
\multirow{3}{*}{Human} & G1      & 1.2124    & 0.0396 & 0.1284 \\
                       & ATLAS   & 1.4008    & 0.0336 & 0.1272 \\
                       & Kinova  & 3.4249 & 0.0491  & 0.0764 \\ \midrule
ATLAS                  & JVRC    & 0.9862    & 0.0329 & 0.1218 \\
JVRC                   & G1      & 1.1174    & 0.0398 & 0.1245 \\
TIAGO                  & ATLAS   & 1.1471    & 0.0455 & 0.1288 \\ 
Kinova    & TIAGo++   &  1.5011   & 0.0372 & 0.0902 \\ \midrule
\end{tabular}%
\caption{\textbf{Adding New Robots.} We assess the generalizability and scalability of our latent space by introducing three new robots—G1, ATLAS, and Kinova. For new robots, we freeze most networks ($E_h, E_X$, and $D_X$) and only train the robot-specific embedding layers ($E_r$ and $D_r$).}
\label{tab:new-robot}
\end{table}

To assess the scalability of our method in incorporating new robots, we demonstrate that minimal additional effort is required—specifically, only a single robot-specific embedding layer needs to be learned, while the rest of the network remains fixed.

Our model is first trained end-to-end on the four primary robots introduced earlier. Once training is complete, we freeze the core networks ($E_h$, $E_X$, and $D_X$) and train only the robot-specific embedding layers ($E_r$ and $D_r$) for the newly added robots—ATLAS, G1, and Kinova Gen3. This approach dramatically reduces the training time, requiring only about 15 minutes, compared to several hours for full retraining from scratch. We evaluate these new robots on both human-to-robot and robot-to-robot motion retargeting tasks, with results presented in Table~\ref{tab:new-robot}. Across all evaluation metrics, the performance of the new robots matches that of the original robots trained end-to-end. These results highlight the generalizability of the learned latent space across different embodiments, making our method highly scalable and adaptable to new platforms with minimal retraining.

\subsubsection{Visual Resemblance}
\begin{figure*}[]
    \centering
    \includegraphics[width=0.98\textwidth]{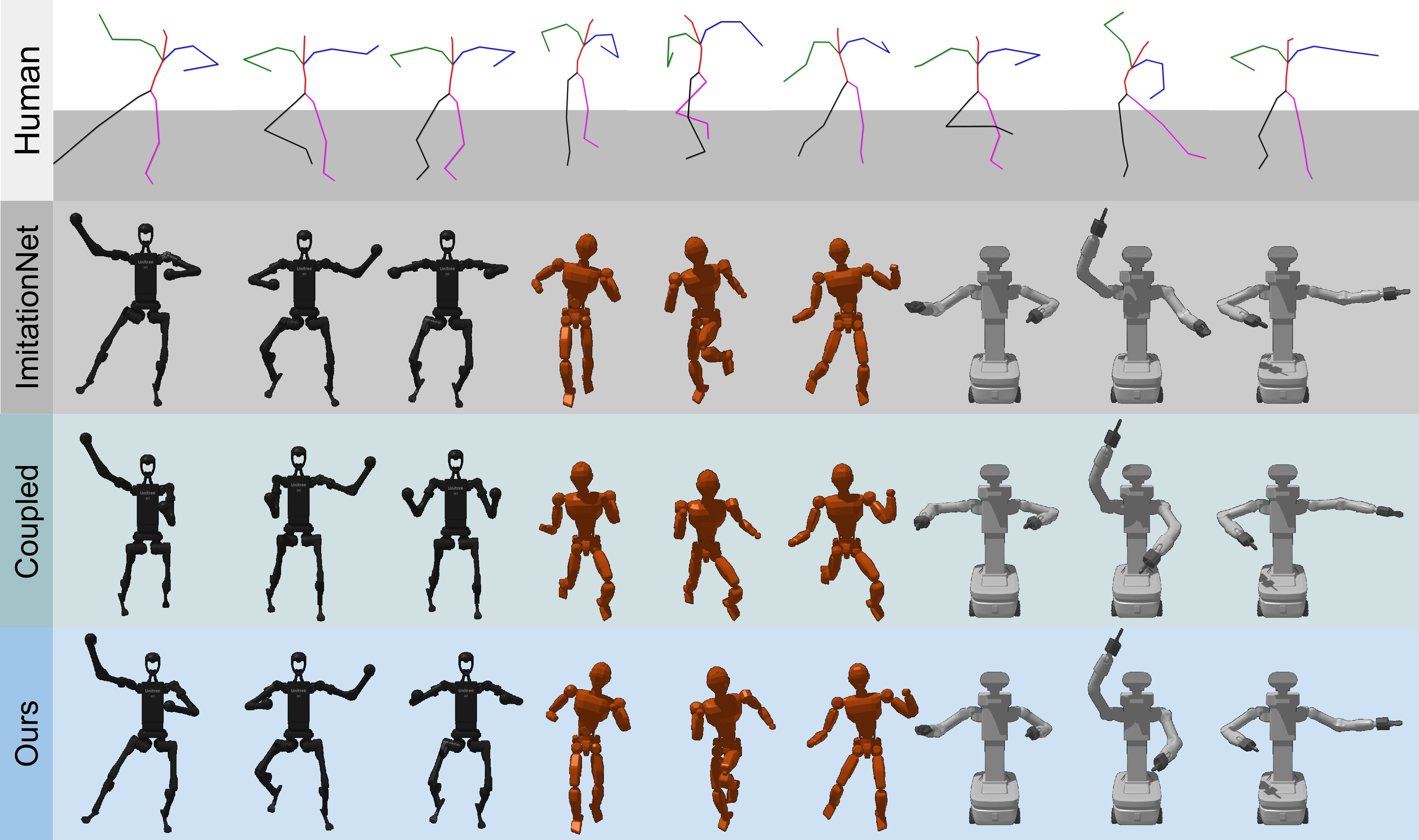}
    \caption{\textbf{Comparison of visual resemblance.} We retarget various dynamic human motions onto different robots (H1, JVRC, and TIAGo, from left to right), and compare different retargeting models. The result shows that both ImitationNet and our method with a decoupled latent space obtain high-quality visual resemblance. Our method trains a single model on all robots, while ImitationNet overfits each robot to a separate model.}
    \label{fig:human-to-robot}
\end{figure*}

\begin{figure*}[]
    \centering
    \includegraphics[width=0.98\textwidth]{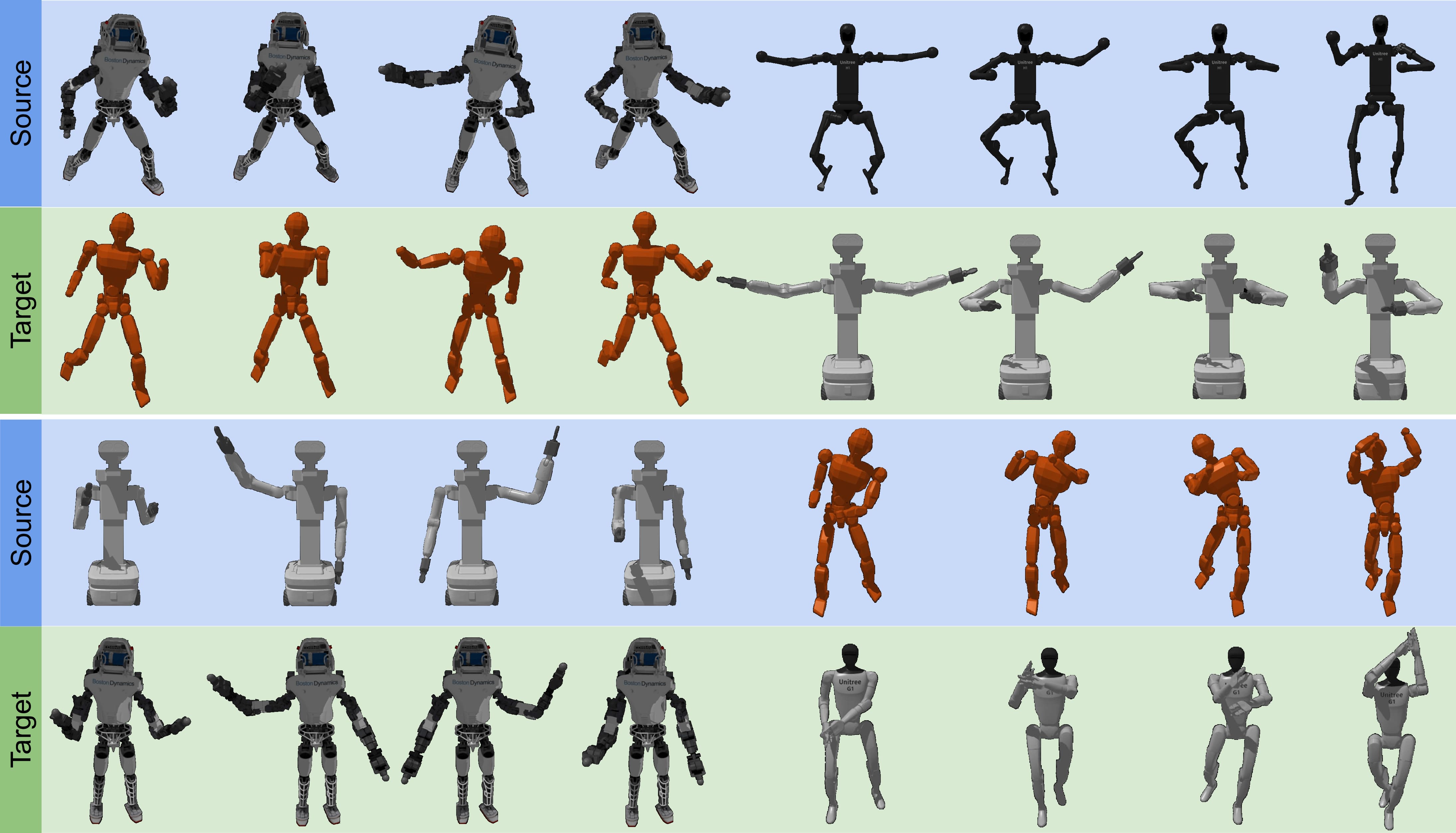}
    \caption{\textbf{Cross-embodiment Motion Retargeting.} We translate motions between any embodiments, and the results showcase the capability of our method, capturing the motion semantics across diverse embodiments.}
    \label{fig:robot-to-robot}
\end{figure*}

A primary goal of motion retargeting is to transfer motions across different embodiments while maintaining their visual fidelity. In Fig. \ref{fig:human-to-robot}, we present a qualitative comparison between our method and the baseline using dynamic, visually expressive motions. The top row depicts the original human motions to be retargeted. ImitationNet delivers high visual similarity by training a dedicated model for each robot. In contrast, our method using a coupled shared latent space struggles to accurately capture leg motions across embodiments. However, when the latent space is decoupled by body segments, our approach matches the visual quality of ImitationNet while retaining the benefit of a shared, generalizable latent space across all robots.

Figure \ref{fig:robot-to-robot} presents additional results demonstrating our method’s ability to retarget motions between different robots. The results show that, despite variations in embodiment, the shared latent representation effectively preserves the visual resemblance of the original motion. Notably, our method achieves high-quality motion transfer across robots with significantly different kinematic structures—for example, from the legged H1 robot to the mobile-based TIAGO, and from TIAGO back to the legged ATLAS robot.

\subsubsection{Latent-Space Robot Control}
\begin{table}[]
\centering
\begin{tabular}{ccccccc}
 & \multicolumn{6}{c}{Distance to Goal (in cm)} \\ \cmidrule(lr){2-7}
                               & TIAGO  & H1     & NAO    & JVRC   & \multicolumn{1}{c}{G1} & \multicolumn{1}{c}{ATLAS} \\ \midrule
c-VAE & 1.14 & 0.44 & 0.13 & 0.45 & 0.42 & 0.56  \\ 
\end{tabular}
\caption{We evaluate the latent-space policy on multiple robots and showcase the performance of goal reaching for each robot.}
\label{tab:laten-space-control}
\end{table}
\begin{figure}
    \centering
    \includegraphics[width=\linewidth]{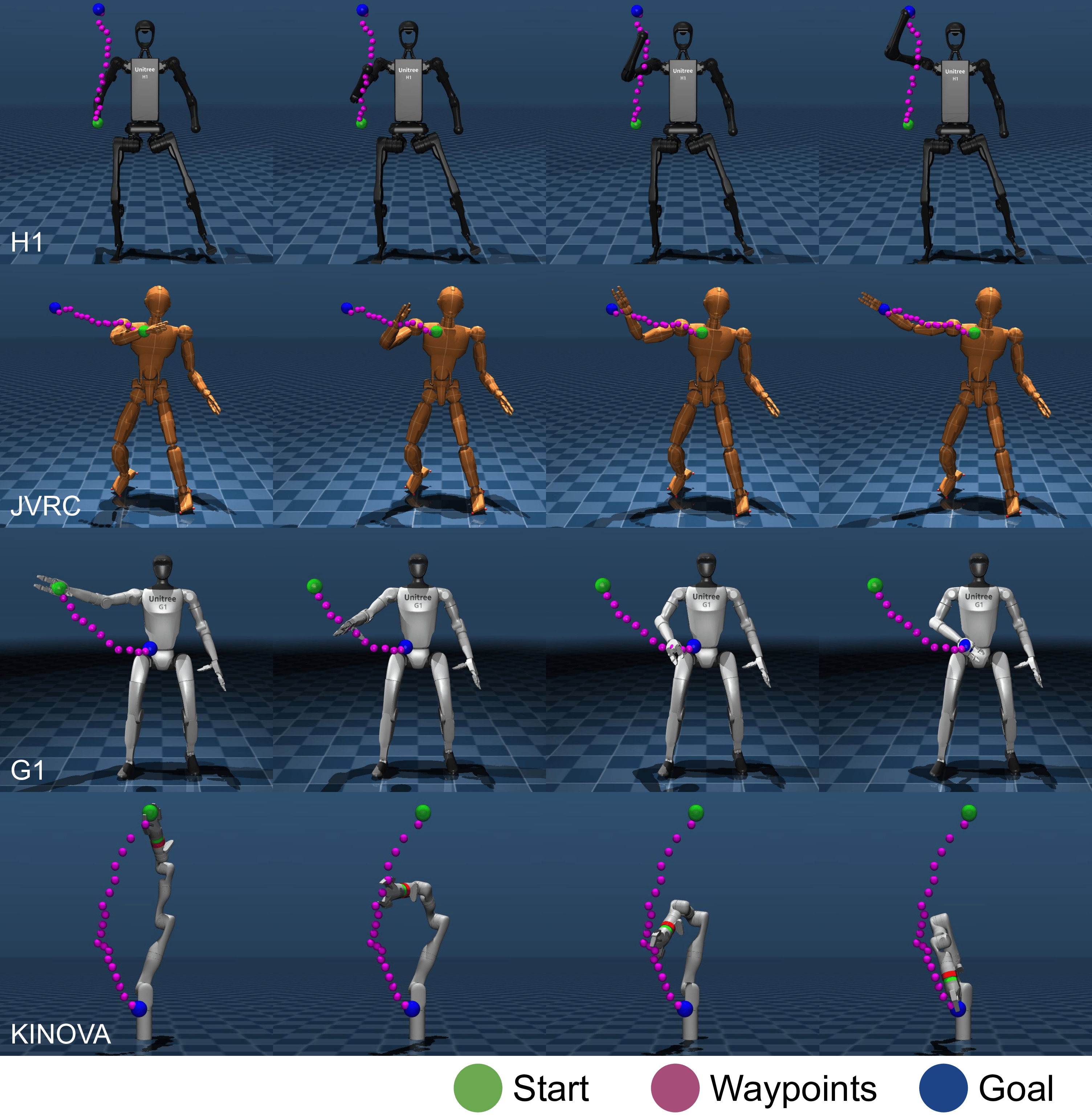}
    \caption{\textbf{Latent-Space Multi-Robot Control.} A single policy controls the latent space to enable each robot to reach arbitrary goal positions (blue) from various starting poses (green). Intermediate waypoints along the generated trajectories are shown in purple.}
    \label{fig:robot_control}
\end{figure}

A key objective of our shared latent space is to enable the learning of control policies that generalize across different robot embodiments. Following the method described in Section~\ref{sec:latent-space-robot-control}, we train a latent-space control policy conditioned on the robot’s end-effector goal position. We evaluate this policy using two criteria: (i) the distance to goal (DTG) metric, which measures the accuracy of reaching the specified EE goal positions, and (ii) computational efficiency. Owing to the decoupled structure of our latent representation, each body segment can be independently controlled. In this study, we focus on learning a policy for controlling the right arm.

To assess the performance of the learned policy, we conduct 1,000 experiments per robot, each with a randomly sampled initial pose and goal EE position. The results, summarized in Table~\ref{tab:laten-space-control}, show that our latent policy enables control of the robotic EE across diverse platforms and achieves sub-centimeter accuracy. Additionally, our latent-space robot control framework is also computationally efficient, supporting control frequencies of approximately 100 Hz.

Figure~\ref{fig:robot_control} visualizes representative trajectories generated by our latent-space control policy across different robots. Each robot starts from a random initial pose, with EE positions indicated by green spheres. The randomly sampled goal positions are shown in blue, and intermediate waypoints—shown in purple—highlight the smoothness of the generated motion trajectories, further demonstrating the effectiveness of our learned controller.

\subsubsection{Motion Editing within Latent Space}
\begin{figure}
    \centering
    \includegraphics[width=\linewidth]{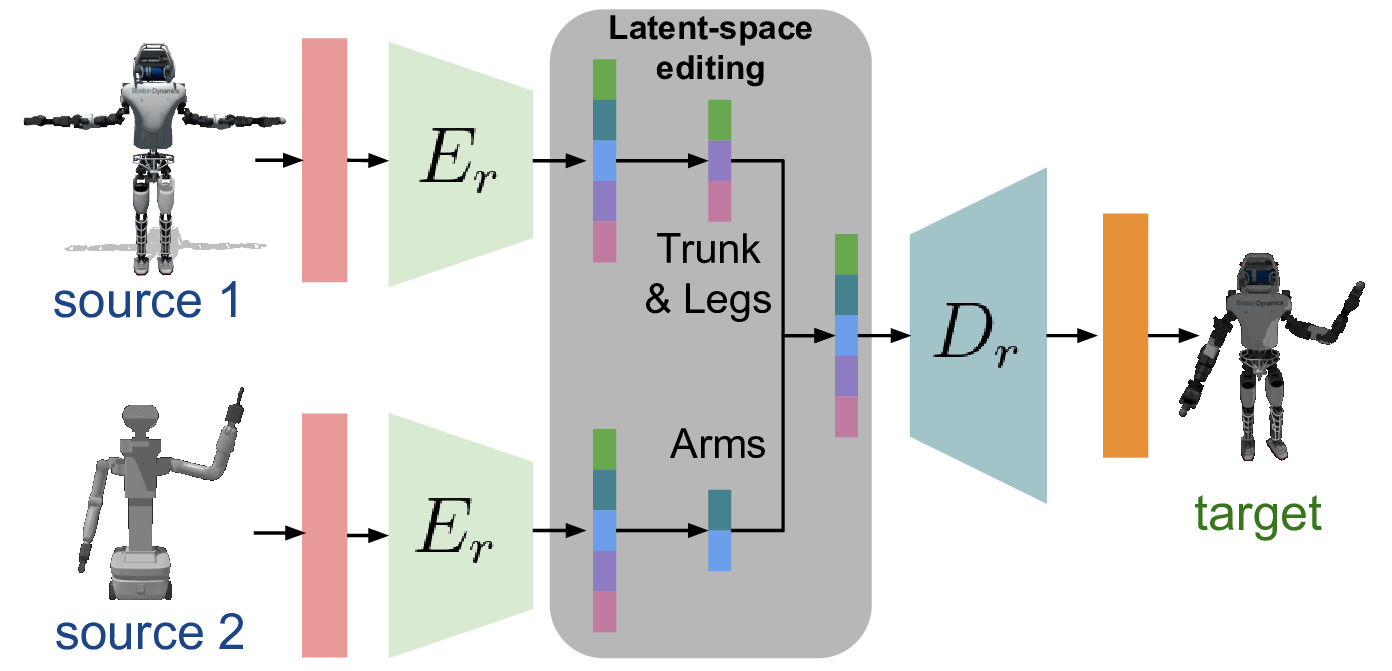}
    \caption{\textbf{Latent-Space Editing.} Our latent space enables motion editing in the latent space, such as composing new motions by combining partial body movements from different sources.}
    \label{fig:latent-space-editing}
\end{figure}
\begin{figure}
    \centering
    \includegraphics[width=\linewidth]{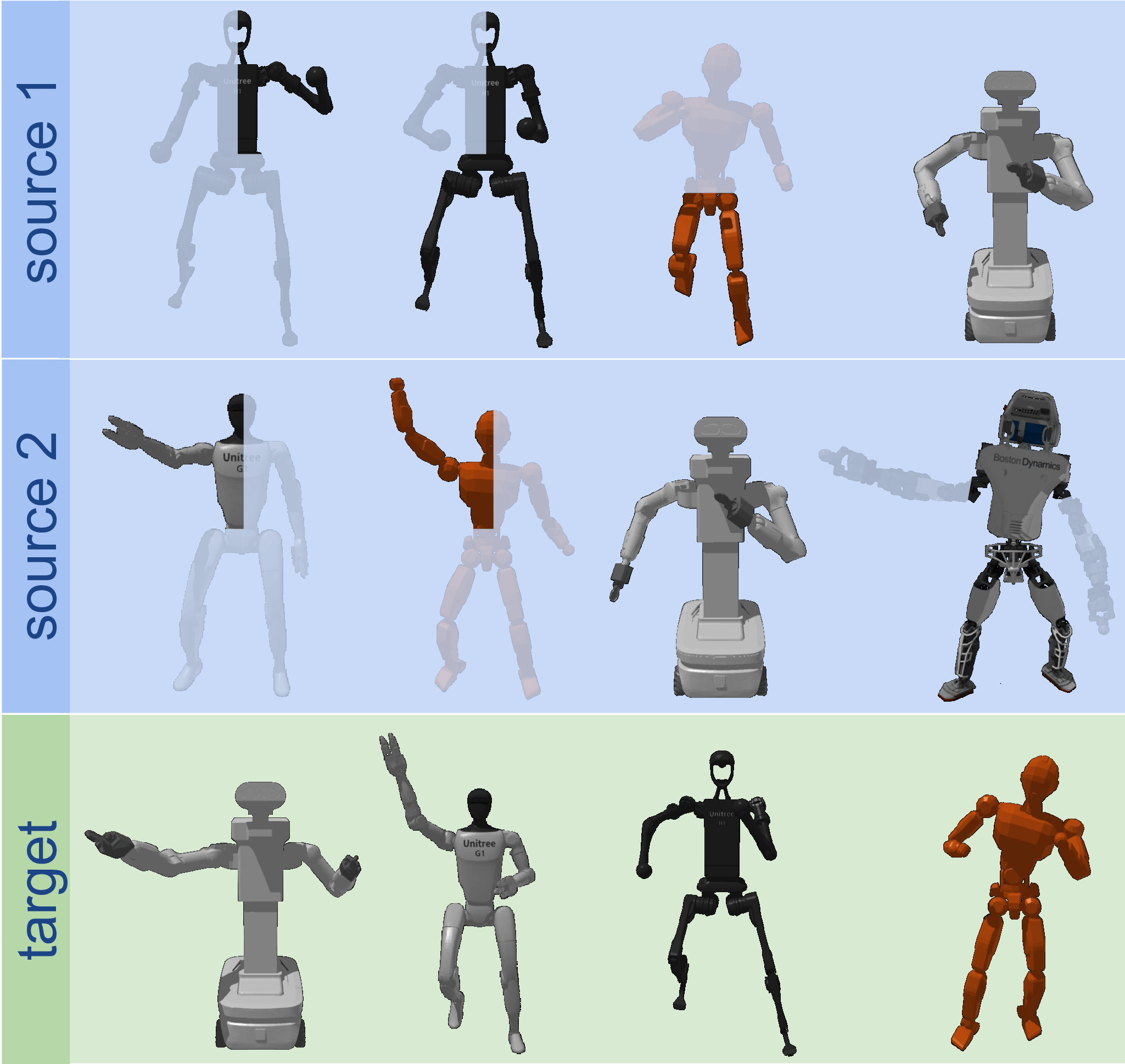}
    \caption{\textbf{Motion Editing.} Our method can synthesize new motions for any target embodiment from different source embodiments by composing their shared latent space.}
    \label{fig:motion-editing}
\end{figure}
Thanks to the decoupled design of our latent space, our method enables a range of versatile applications. One example is independent control of individual body segments, as demonstrated in Figure \ref{fig:robot_control}. Additionally, the learned latent space supports motion editing capabilities, such as composing new motions by combining partial movements from different sources. Figure \ref{fig:latent-space-editing} illustrates this functionality: in the shown example, we synthesize a new ATLAS motion by stitching together the arm motion from TIAGO with the trunk and leg motion from ATLAS—seamlessly blended within the latent space. Figure \ref{fig:motion-editing} showcases additional examples of motion synthesis, where target-domain motions are generated by combining source-domain motions from different embodiments. The high-quality visual consistency of the composed motions highlights the effectiveness of our cross-domain motion retargeting and the strong decoupling capabilities of the learned shared latent space.

\subsubsection{Ablation Study}
\begin{table}[]
\centering

\begin{tabular}{ccccccc}
& \multicolumn{3}{c}{Human-to-TIAGo++} & \multicolumn{3}{c}{Human-to-JVRC}\\ \cmidrule(lr){2-4} \cmidrule(lr){5-7}
$w, \lambda_c$ & RS & NDS & NVS & RS & NDS & NVS\\ \midrule
 $1.0, 5.0$ & 4.2224 & 0.0487 & 0.1077  & 1.6395 &   0.0367 & 0.0919\\
$\mathbf{1.0}, \mathbf{10}$ & 3.8293 &  0.0401  &   0.1071  & 1.4631  & 0.0288  & 0.0862\\
$1.0, 15$  &  3.6749 &  0.0482  & 0.1111 & 1.3970 & 0.0323  & 0.0942 \\
$0.5, 10$  &  3.0738  & 0.0460  &  0.1020  &  1.1512 &  0.0349  & 0.0880 \\
$1.5, 10$ & 3.2215 &  0.0495  &  0.1100 &  1.1814  &    0.0306  & 0.1010        
\end{tabular}
\caption{\textbf{Ablation Study on Hyperparameter}. $w$ and $\lambda_c$ are essential parameters for the similarity metric in Eq. \ref{eq:similarity} and objective function in Eq. \ref{eq:total_loss}. Bold indicates the selected parameters.}
\label{tab:ablation}
\end{table}
To analyze the effect of hyperparameters on the training process, we conduct an ablation study on two key parameters, $w$ and $\lambda_c$, defined in Eq.~\ref{eq:similarity} and Eq.~\ref{eq:total_loss}, respectively. The parameter $w$ controls the balance between limb rotations and hand positions in the similarity metric, while $\lambda_c$ determines the weight of the contrastive loss in the overall objective. Table~\ref{tab:ablation} reports the motion retargeting performance on the TIAGo and JVRC robots under different parameter settings. Our selected configuration, $w=1.0$ and $\lambda_c=10$, achieves a strong trade-off between visual fidelity and end-effector controllability.

\begin{figure*}[]
    \centering
    \includegraphics[width=0.98\textwidth]{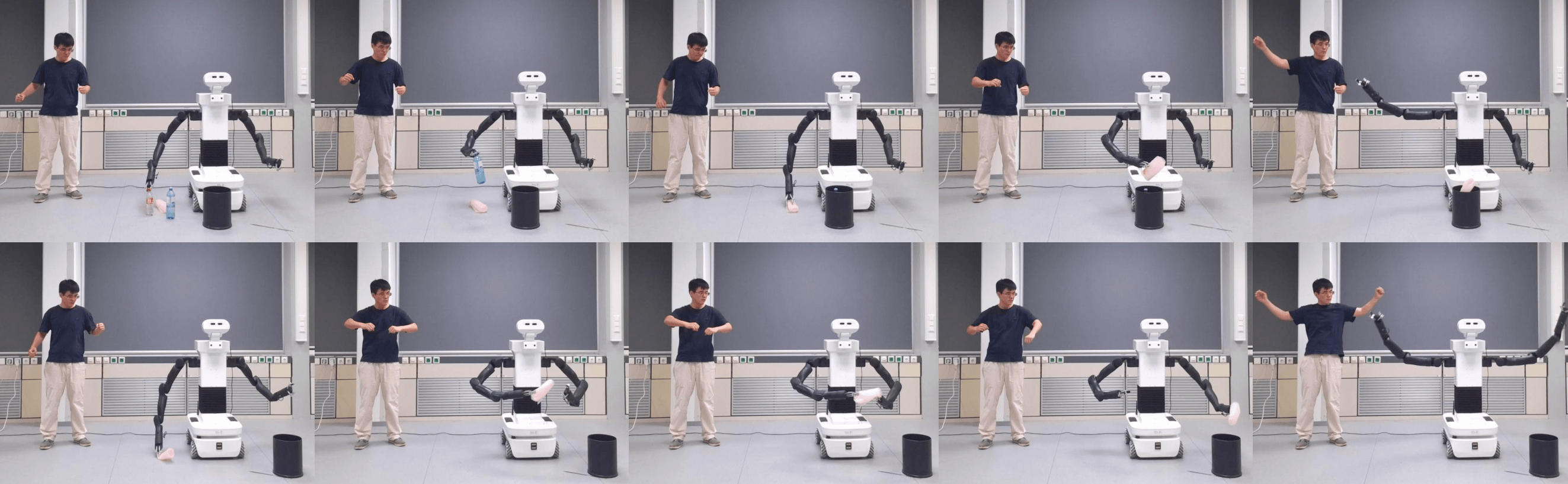}
    \caption{\textbf{Teleoperation from an RGB camera.} The dual-arm TIAGo-SEA is teleoperated in real time to perform object pick-and-place (top) and bimanual object transfer from the right hand to the left hand (bottom).}  
    \label{fig:teleopt}
\end{figure*}

\begin{figure*}
    \centering
    \includegraphics[width=0.98\textwidth]{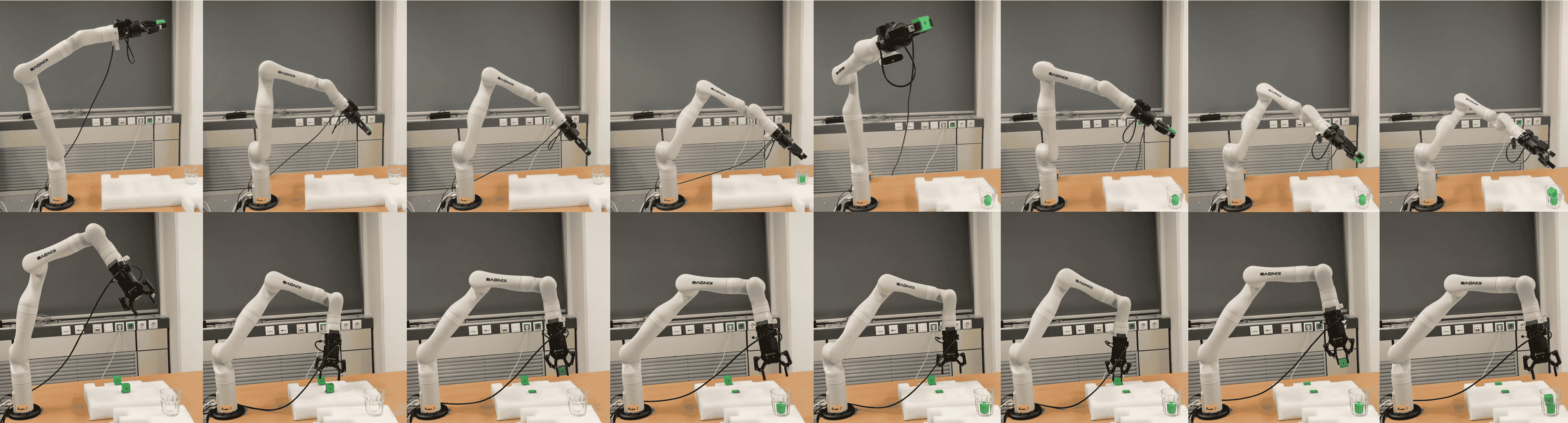}
    \caption{\textbf{Robot Control via Latent Space.} By operating in the latent space, the Kinova robot performs tasks such as reaching target positions and executing pick-and-place.}  
    \label{fig:kinova-control}
\end{figure*}

\begin{figure}[]
    \centering
    \includegraphics[width=0.98\linewidth]{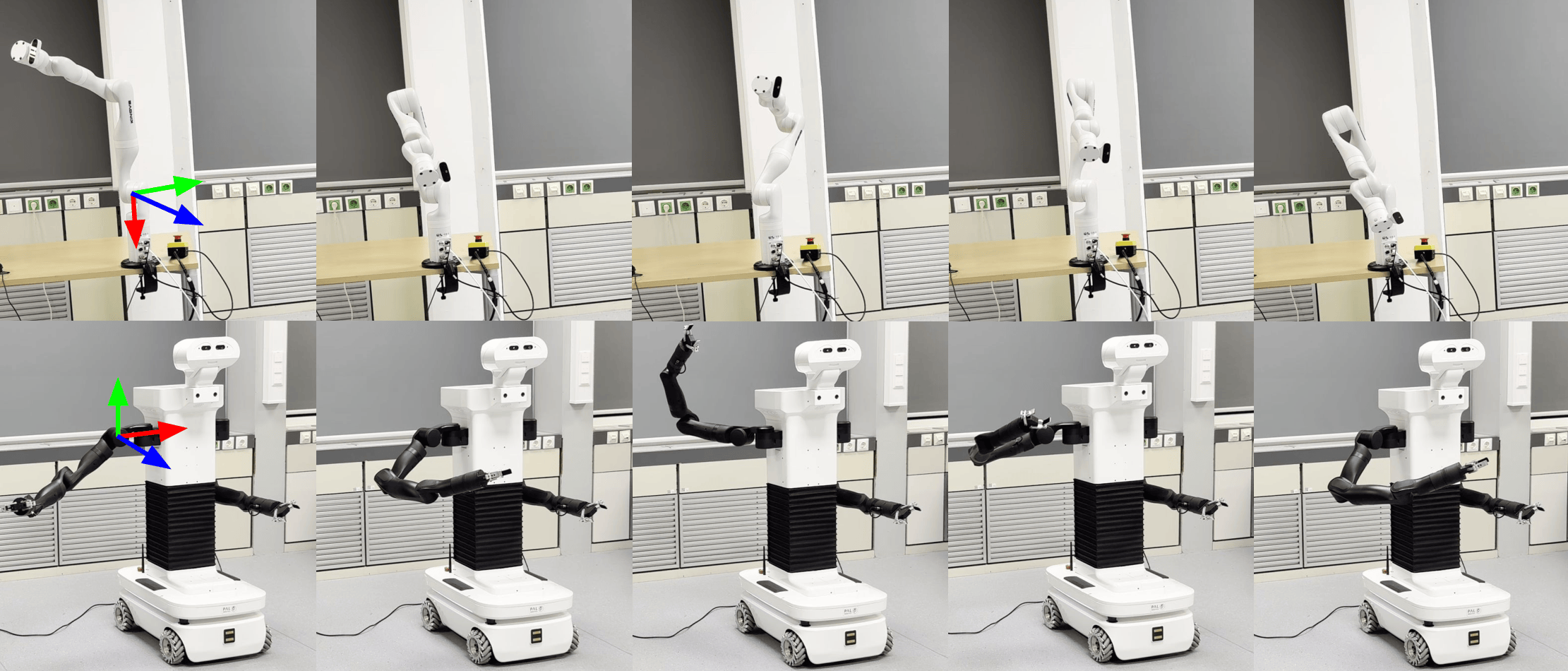}
    \caption{\textbf{Robot-to-Robot Retargeting.} The TIAGo-SEA’s right arm follows the motion of the Kinova robot, after applying a transformation to better align the workspaces of the two robots.}
    \label{fig:kinova-tiago}
\end{figure}
\subsubsection{Experiments on Hardware}

We test our method on two different robot platforms: the TIAGo-SEA and Kinova Gen3 robot. To fully evaluate the capabilities of our method, we conduct different experiments: human-to-robot and robot-to-robot motion retargeting, and robot control through our latent space.

Figure \ref{fig:teleopt} shows real-time robot teleoperation using only an RGB camera with our method. The dual-armed TIAGo-SEA robot is teleoperated to perform object pick-and-place by throwing items into a trash bin. We also demonstrate bimanual object transfer, highlighting the utility of dual-arm teleoperation enabled by our approach.

In Figure \ref{fig:kinova-tiago}, the motion of the Kinova arm is transferred to the TIAGo-SEA’s right arm. Since the Kinova is mounted in a tabletop setting, we transform its base link to correspond to the right shoulder, thereby aligning the workspaces of the two robots.

Furthermore, Figure \ref{fig:kinova-control} demonstrates our method’s ability to control robots directly in the shared latent space. The Kinova robot is controlled to pick up objects and drop them into a glass. Successful execution of these tasks requires precise end-effector control, which highlights the effectiveness of our method.

\subsection{Discussion and Future Work}
In this work, we proposed an unsupervised learning approach to construct a cross-embodiment latent space in which semantically similar motions are closely aligned, regardless of the source embodiment. Leveraging this shared representation, we introduced a c-VAE framework to learn goal-conditioned latent control policies from only human data, and the policies can be directly deployed across diverse robotic platforms.

Currently, our motion retargeting is trained on the HumanML3D dataset, which contains human motions in the SMPL model. Since the SMPL model does not capture hand movements, it considers the hands as part of the forearms. As a result, due to the missing hand motion in the SMPL, hand motion retargeting is not handled in this paper, limiting applications that require fine-grained teleoperation. In future work, we plan to address this limitation by using existing (or collecting) hand datasets to incorporate human hand imitation, enhancing the robot's latent control policy to support a wide range of manipulation tasks.

\begin{figure*}
    \centering
    \includegraphics[width=0.98\linewidth]{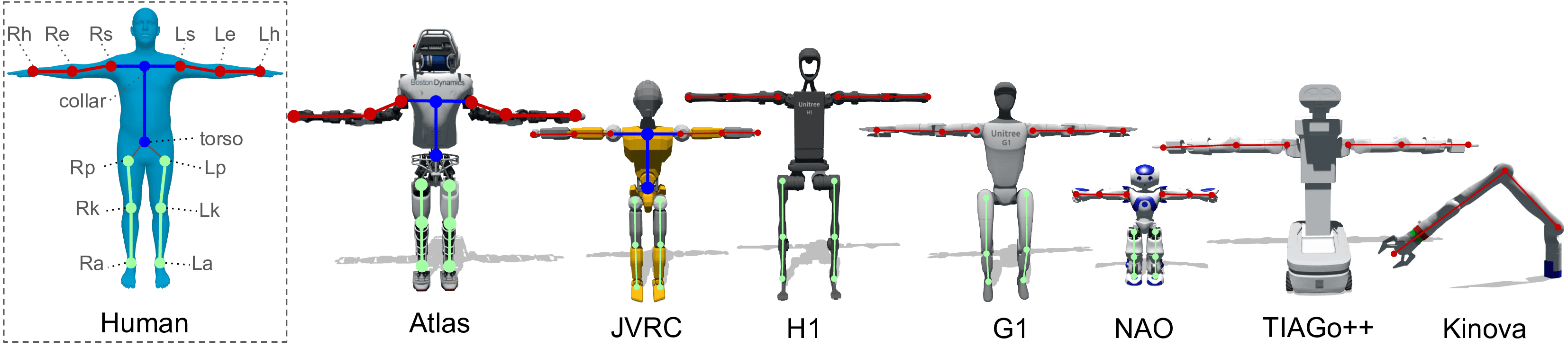}
    \caption{Mapping among different embodiments.}
    \label{fig:mapping}
\end{figure*}
\begin{figure*}
    \centering
    \includegraphics[width=0.98\linewidth]{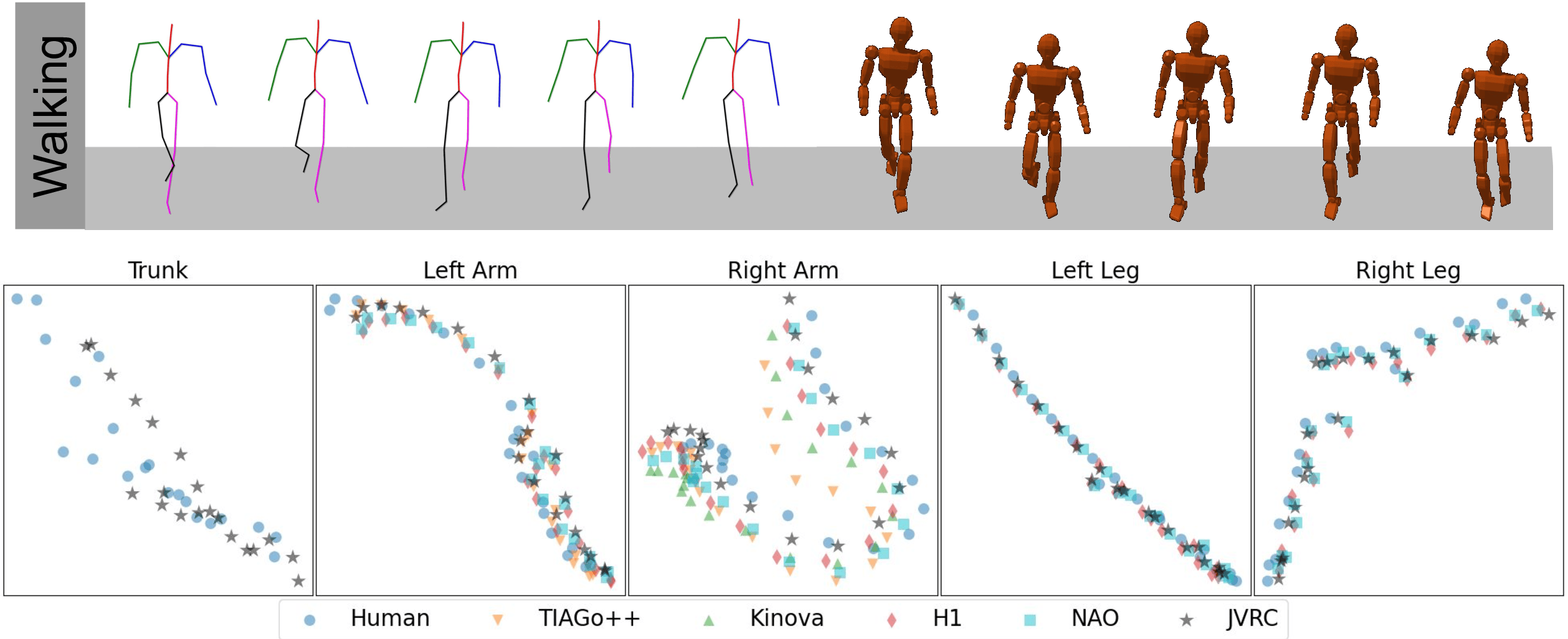}
    \caption{\textbf{Visualization of Latent Space.} After dimension reduction with PCA, we visualize the projected latent space trajectories of the retargeted walking motion from multiple robots.}
    \label{fig:pca}
\end{figure*}

\section{Conclusions}
Cross-embodiment control is a critical capability for enabling scalable and flexible multi-robot systems. In this work, we introduced a two-stage framework to address this challenge: (1) learning a shared latent space that encodes semantically similar motions across diverse embodiments, and (2) training control policies within this space that can be deployed on a wide range of robots without further adaptation.

In the first stage, we leverage contrastive learning to build a generalizable and scalable latent representation across multiple robot and human embodiments. We validate this latent space through extensive quantitative and qualitative experiments on both human-to-robot and robot-to-robot motion retargeting. When adding new robots to our shared latent space, only a lightweight robot-specific embedding layer needs to be trained, while the rest of the model remains frozen. Despite this minimal adaptation, newly added robots achieve performance comparable to those trained end-to-end, demonstrating the generalizability of the learned latent space. As a result, trained control policies can be directly deployed on new robots, significantly enhancing the efficiency of multi-robot system management.

In the second stage, we introduce a goal-conditioned control method in the latent space using a c-VAE architecture. This policy is trained exclusively on human motions and can be directly deployed across multiple robots. Our experiments show that the learned policy achieves accurate end-effector control, achieving sub-centimeter errors across different platforms.

Overall, our method demonstrates a promising path toward unified, scalable, and adaptable control across diverse robot embodiments.

\begin{appendices}

\section{Robots with different embodiments}
\label{appendix:robots}

\begin{table}[]
\centering
\caption{Human body parts that robots can mimic}
\begin{tabular}{l|c|c|c|c}
Robot & Arms & Trunk & Legs & Arm-Ratio \\ \hline
ATLAS & \cmark & \cmark &  \cmark & 1.00 \\
H1 & \cmark & \xmark & \cmark & 0.90\\
G1 & \cmark & \xmark & \cmark & 1.43 \\
JVRC & \cmark & \cmark & \cmark & 0.78\\
NAO & \cmark & \xmark & \cmark & 0.53 \\
TIAGo++ & \cmark & \xmark & \xmark  & 1.16 \\
Kinova & \cmark &  \xmark & \xmark & 1.59\\
\end{tabular}

\label{table:robots}
\end{table}
We evaluate our method on six robots that span a wide range of morphologies, from mobile-based bimanual platforms to fully bipedal humanoids. Figure \ref{fig:mapping} visualizes the kinematic mappings between different embodiments, while Table \ref{table:robots} summarizes the capabilities of each robot. For example, the ATLAS robot has fully actuated arms, legs, and a trunk, whereas the TIAGo robot is limited to arm movements only. The final column in the table reports the arm ratio—defined as the length ratio between the forearm and upper arm—for each robot. These ratios vary significantly, ranging from 0.53 to 1.43, reflecting the morphological diversity of the platforms and highlighting the inherent challenges of developing a unified multi-robot control framework.

\section{Latent Space Visualization}

To better understand the structure of the shared latent space constructed across multiple embodiments, we visualize the projected latent trajectories using principal component analysis (PCA), as shown in Fig. \ref{fig:pca}. An example walking motion (top) is first retargeted to several robots, such as the JVRC. The resulting robot motions are then encoded into the shared latent space, and their projected trajectories are plotted in Fig. \ref{fig:pca} (bottom). The visualization demonstrates the alignment and consistency of the motion representations across different embodiments within the shared latent space.

\end{appendices}


\bibliographystyle{IEEEtran}
\bibliography{references}

\newpage

\vfill

\end{document}